\pdfoutput=1

\documentclass[11pt]{article}
\usepackage{colortbl}
\usepackage{pgfplots}
\usepackage{emnlp2021}

\usepackage{times}
\usepackage{latexsym}

\usepackage[T1]{fontenc}

\usepackage[utf8]{inputenc}

\usepackage{microtype}

\usepackage{times}
\usepackage{amsmath}
\usepackage{algorithm}
\usepackage{algpseudocode}
\usepackage{diagbox}
\usepackage{latexsym}
\usepackage{enumitem}
\usepackage{siunitx}

\usepackage{helvet}  
\usepackage{courier}  
\usepackage{url}  
\usepackage{graphicx}  
\usepackage{courier}
\usepackage{amsmath}
\usepackage[normalem]{ulem}
\useunder{\uline}{\ul}{}
\usepackage{graphicx}
\usepackage{multirow}
\usepackage{mathbbol}
\usepackage{caption}
\usepackage{CJKutf8}
\usepackage{verbatim}
\usepackage{float}
\usepackage{booktabs}
\usepackage{array}
\usepackage{times}
\usepackage{csquotes}
\usepackage{latexsym}
\usepackage{tablefootnote}
\usepackage[normalem]{ulem}
\usepackage{todonotes}
\usepackage{subcaption}
\usepackage{tikz}

\makeatletter
\def\BState{\State\hskip-\ALG@thistlm}
\makeatother

\title{\textbf{HypoGen}: Hyperbole Generation with Commonsense and Counterfactual Knowledge}

\author{Yufei Tian, Arvind krishna Sridhar, and  Nanyun Peng \\
  Computer Science Department, University of California, Los Angeles \\
  {\tt \{yufeit, arvindkrishna97, violetpeng\}@cs.ucla.edu} \\}

\begin{document}
\maketitle
\begin{abstract}
A hyperbole is an intentional and creative exaggeration not to be taken literally. Despite its ubiquity in daily life, the computational explorations of hyperboles are scarce. In this paper, we tackle the under-explored and challenging task: sentence-level hyperbole generation. 
We start with a representative syntactic pattern for intensification and systematically study the semantic (commonsense and counterfactual) relationships between each component in such hyperboles. Next, we leverage the COMeT and reverse COMeT models to do commonsense and counterfactual inference. We then 
generate multiple hyperbole candidates based on our findings from the pattern, and train neural classifiers to rank and select high-quality hyperboles. 
Automatic and human evaluations show that our generation method is able to generate hyperboles creatively with high success rate and intensity scores.
\end{abstract}

\section{Introduction}\label{sec:intro}
\begin{figure}[t!]
\centering
\includegraphics[width=0.5\textwidth]{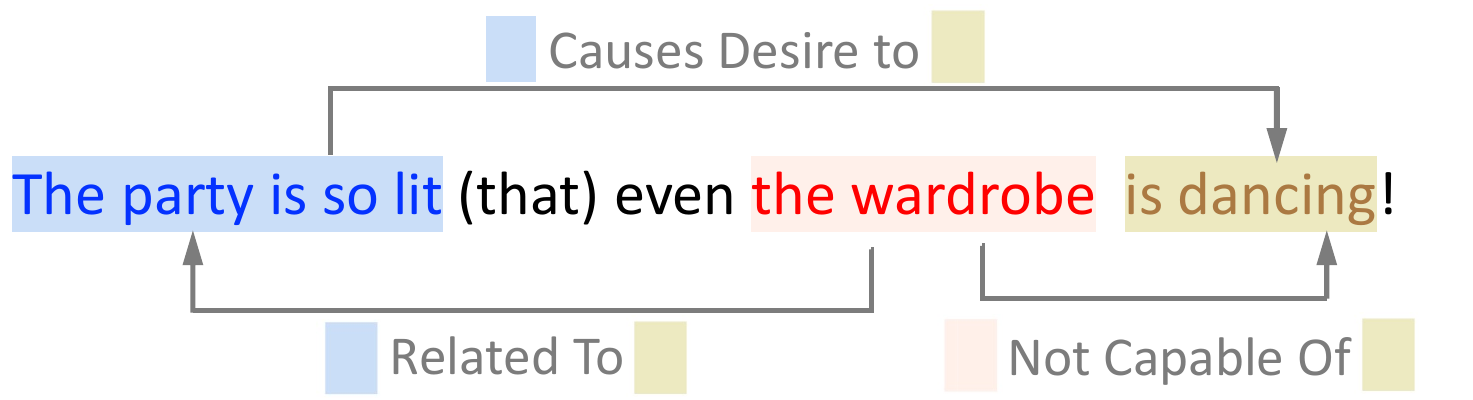}
\caption{An illustration of the commonsense and counterfactual relationships within a \textit{clause or sentence level} hyperbole. The {\color{blue} input prompt ($\mathbf{A}$)}, {\color{red} subject in the clause ($\mathbf{B}$)}, {\color[HTML]{998403} predicate in the clause ($\mathbf{C}$)}, and {\color[HTML]{656565} the relationships between them} are colored in blue, red, brown and grey. In this example, that \textit{`the party is lit'} causes the desire to \textit{`dance'}. In addition, \textit{`the wardrobe'} is related to \textit{`the party'}, and is not capable of \textit{`dancing'}.} 
\label{Table:example}
\vspace{-1.5em}
\end{figure}

Hyperboles invoke the use of exaggeration as a rhetorical device or figure of speech. It is interactive, amusing, and is the second most common among all tropes of figurative language, only after metaphors \cite{Kreuz1996}. 
 By definition, a hyperbolic expression exceeds the credible limits of fact in the given context, whereas a literal expression agrees with the extralinguistic facts in the given context \cite{claridge2010hyperbole}. For example in Figure \ref{Table:example}, \textit{``The party is so lit even the wardrobe is dancing!''} is considered as a hyperbole because making a lifeless object to dance is impossible; it is an intentional and creative way of exaggerating how lit the party is, and is not meant to be taken literally. In contrast, \textit{``The party is so lit (that) even my introvert friend has a good time!''} is considered literal, because letting introvert people have a good time is realistic and hence not an overstatement.

Despite its abundance, identifying and generating hyperboles remain under-explored. Compared to the many efforts on other figurative languages such as puns, sarcasms, metaphors and similes \cite{he2019pun, chakrabarty2020r, su2020deepmet, yu2019avoid, chakrabarty2020generating}, the exploration of hyperboles is still in the infancy stage: NLP researchers have just started to look at automatic hyperbole detection \cite{troiano2018computational, kong2020empirical}. According to \citet{claridge2010hyperbole}, hyperboles are divided into two categories: those at the \textit{word or phrase level} and those at the \textit{clause or sentence level}. The former is less creative because it is easily achievable via lexicon substitution \citep{norrick2012semantics}. For example, replacing most time durations with `a millisecond'
will make noncreative exaggerations to emphasize something is fast, without needing to understand the context.

In this work, we target at generating the more challenging type of hyperboles, i.e. \textbf{clause or sentence level hyperboles}.
According to \citet{mccarthy2004there}, clause-level hyperboles consist of counterfactuality and syntactic support. 
Inspired by the linguistic theory that `\textit{so} + \textit{adj/adv} + \textit{(that)} + \textit{a declarative clause}' is a significant pattern with both prototypical syntactic and semantic function as overstatement \cite{backlund1973collocation, lorenz2002really}, we leverage the \textit{so...that} pattern, where `that' is omittable, as a starting point to analyze and generate hyperboles. 

\citet{claridge2010hyperbole} state that within the \textit{so...that} pattern, the sentence serves as a result of the prompt ($\mathbf{A}$) and that the sentence itself creates impossible worlds. 
Inspired by this, we systematically investigate the semantic (commonsense or counterfactual) relationships between the components within the \textit{so...that} pattern. Specifically, we partition each sentence into three parts: {\color{blue}the literal prompt ($\mathbf{A}$)}, {\color{red}the subject of the clause ($\mathbf{B}$)}, and {\color[HTML]{998403}the predicate of the clause ($\mathbf{C}$)}, as illustrated in Figure \ref{Table:example}, and conduct detailed annotation and analysis. We discover that 6 semantic relations among $\mathbf{A}$, $\mathbf{B}$, and $\mathbf{C}$ account for over 95\% of all hyperboles with the \textit{so...that} pattern. This indicates that if a generation model can cover these 6 relationships, it is able to generate almost all hyperboles with such pattern.

With the annotated relationships as background knowledge, we build a hyperbole generation model that takes a literal prompt ($\mathbf{A}$) as input and outputs a hyperbole clause ($\mathbf{B}$ and $\mathbf{C}$ combined). To this end, we train a reverse COMeT model to generate commonsense and counterfactual phrases along with the COMeT model \cite{bosselut2019comet}, and rank the generated candidates with a hyperbole identifier.
Finally, we break the restrictions of the \textit{so...that} pattern, and generate hyperboles with diverse syntactic structures using a syntactically controlled paraphrase model. To the best of our knowledge, we are the first to analyze the relations of the logical components within hyperboles, and the first to automatically generate hyperboles. We summarize our contributions as follow: 
\begin{itemize}[leftmargin=0pt]
    \item We create an English hyperbole dataset from the online discussion forum, Reddit, 
    and analyze hyperboles in the \textit{so...that} pattern to understand the commonsense and counterfactual relationships between each component within such pattern. Our analysis discover that 6 major relations cover 95\% of all occurrences. This provide guidelines for us to design models for automatic hyperbole generation. (Details can be found in Section \ref{section:data}) 
    \item Based on the analysis, we propose \textbf{HypoGen}, a hyperbole generation model that takes a literal prompt as input, and generate hyperbole sentences. 
    Automatic and human evaluations show that our best model \textbf{HypoGen}$_{Spec}$ is able to generate high-quality hyperboles with high success rate.\footnote{Our code and data are  available at \url{https://github.com/NinaTian98369/HypoGen}} (Details can be found in Section \ref{sec:model})
    \item We further propose to apply syntactically controlled paraphrase generation model to break the \textit{so...that} pattern and generate creative hyperboles with diverse syntactic structures.
    
\end{itemize}


\section{Task Definition}
Given an input prompt ($\mathbf{A}$), we aim to generate clause or sentence level hyperboles by completing that clause. For example, if the input is \textit{`the party is lit'}, our task is to generate \textit{`the wardrobe'} (a subject $\mathbf{B}$) and \textit{`is dancing'} (a predicate $\mathbf{C}$) to make the full sentence (\textit{`the party is so lit that even the wardrobe is dancing'}) a hyperbole. 


\section{Data Collection and Analysis} \label{section:data}
Section \ref{sec:data1} introduces how we collect hyperboles and non-hyperboles sentences from Reddit. In Section \ref{sec:rel-anno}, we describe the procedure for a detailed second-round annotation: sensical relationship annotation for hyperboles with the \textit{so...that} pattern.

\subsection{Collection of Hyperboles}\label{sec:data1}
Considering their ubiquity in people's everyday conversation, we collect hyperboles from online discussion forums. We first crawl thousands of sentences from Reddit that contain different patterns or adverb keywords (phrases) that are potential hyperboles, such as \textit{I swear}, \textit{literally}, and \textit{so...that} \cite{mora2009all}. Table \ref{Table:reddit} illustrates how the retrieved sentences containing such keywords can be both hyperboles (positive examples) and literal (negative examples) sentences. Thus, we instruct human annotators to decide if a given sentence is hyperbole or not. In total, 3,300 sentences are annotated and each sentence is annotated by at least three annotators. The worker agreement with aggregate, or "Wawa", which measures the average number of times that the rators' response agree with the aggregate answer
, is 0.72. 

\begin{table}[t!]
\small
\centering
\begin{tabular}{|l|l|}
\hline
Literal & \begin{tabular}[c]{@{}l@{}}Postgraduate \textbf{literally} refers to any degree\\ after an undergraduate degree.\end{tabular} \\ \hline
Hyperbole & \begin{tabular}[c]{@{}l@{}}My boyfriend was so hungry, he \textbf{literally}\\ swallowed his plate.\end{tabular}              \\ \hline
Literal & \begin{tabular}[c]{@{}l@{}}\textbf{I swear} to God I don't know how that cat\\ got there! \end{tabular}                 \\ \hline
Hyperbole & \begin{tabular}[c]{@{}l@{}} \textbf{I swear} to Jeebus I will burn this building\\ to the ground!\end{tabular}                        \\ \hline
\end{tabular}
\caption{Examples of retrieved sentences from Reddit that contain keywords `literally' and `I swear'. Whether these sentences are hyperbole or literal depends on the semantic meaning, not the existence of such keywords.}
\label{Table:reddit}
\vspace{-1.5em}
\end{table}
\begin{table*}[t!]
\centering
\includegraphics[width=0.9\textwidth]{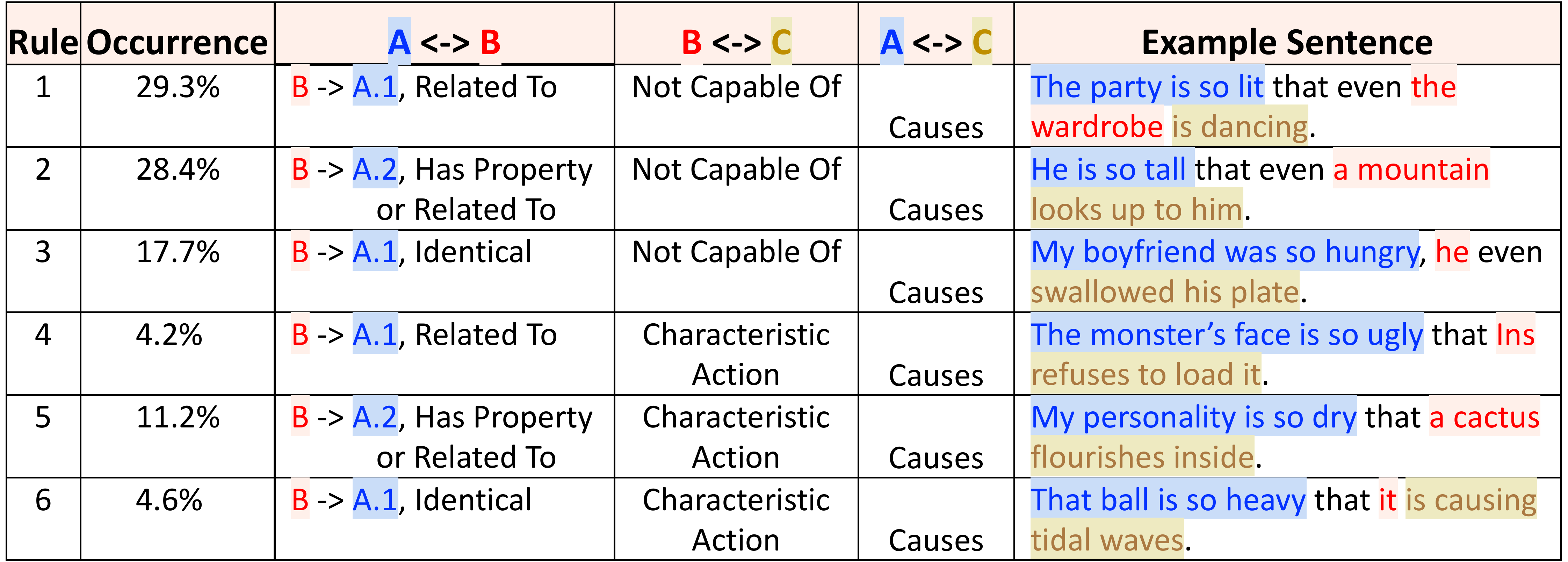}
\caption{Our annotation results: we identify six sensical relations for the \textit{so...that} pattern. We list the percentage of occurrences, names of relation between $\mathbf{AB}$, $\mathbf{AC}$, $\mathbf{BC}$, and example sentences. Here $\mathbf{A.1}$ and $\mathbf{A.2}$ stand for the subject of $\mathbf{A}$ and the head word modified by \textit{so}. }
\label{Table:6-rels}
\vspace{-1.0em}
\end{table*}

We call our collected data HYPO-Red. HYPO-Red is valuable because both negative and positive samples contain such keywords, meaning that an identification model must avoid the superficial patterns and focus on the crux of hyperbole: 1) going to extreme and 2) counterfactuality not meant to be taken literally. Using our collected data, we trained a hyperbole identifier by finetuning BERT. Details can be found in Section \ref{sec:exp-dectect}.

\subsection{Relationship Annotation}\label{sec:rel-anno}

\paragraph{The \textit{so...that} Pattern} We already know that clause-level hyperboles include counterfactuality and syntactic support. Moreover, the content clauses always express the resultant meaning of the prompts (e.g.,  `want to dance' is the result of `the party is lit') and that the clause itself creates impossible worlds (e.g., `wardrobe is dancing' creates an impossible world) \cite{claridge2010hyperbole}. However, those observations are not concrete enough for a systematical exploration of complicated hyperboles and hyperbole generation. To uncover the underlying sensical (commonsense and counterfactual) relationships of hyperboles, we study the \textit{so...that} pattern because it is both representative and easy to spot using keywords. Specifically, we randomly collected 500 hyperboles that contain either \textit{so...that} and \textit{so...even}, and then partition the pattern into three components:  {\color{blue}the literal prompt ($\mathbf{A}$)}, {\color{red}the subject in the clause ($\mathbf{B}$)} and {\color[HTML]{998403}the predicate (verbal phrase) in the clause ($\mathbf{C}$)}. We then instruct six annotators to annotate these 500 hyperboles. 

\paragraph{Annotation Procedure} We provide the annotators with a few seed options present in linguistic papers (such as $\mathbf{C}$ as the result of the $\mathbf{A}$). The annotators are asked to independently label the relationships within a sentence, i.e., between $\mathbf{AB}$, $\mathbf{BC}$, and $\mathbf{CA}$.  All annotators receive detailed instructions about how to react if they find a new sensical relationship or none of the seed options fit. Each sentence is annotated by three people.

\paragraph{Annotation Results} We find that 6 sensical relations account for over 95\% of all occurrences. We report their percentage of occurrences, names for each relation, and example sentences in Table \ref{Table:6-rels}. First, we discover that $\mathbf{C}$ is always the result of $\mathbf{A}$. Next, the interaction of $\mathbf{B}$ and $\mathbf{C}$ creates counterfactuality \citep{claridge2010hyperbole}. Either $\mathbf{B}$ is not capable of conducting the action of $\mathbf{C}$ (rule 1-3), or $\mathbf{C}$ is one of $\mathbf{B}$'s characteristic actions, but surely unrealistic given the context of $\mathbf{A}$ (rule 4-6). For instance, for rule 5, `{\color{red} a cactus}' grows in dry area and `{\color[HTML]{998403}flourish}' is one of its characteristic actions. However, a cactus \textit{cannot} grow inside one's mind. Given the context of `my personality is dry', that `a cactus flourishes inside' is unrealistic.\footnote{Occasionally, $\mathbf{C}$ may also be the \textit{inverse} characteristic action of $\mathbf{B}$, depending on the context of $\mathbf{A}$ (see the example sentence of rule 4).}

\begin{figure*}[t!]
\centering
\includegraphics[width=0.8\textwidth]{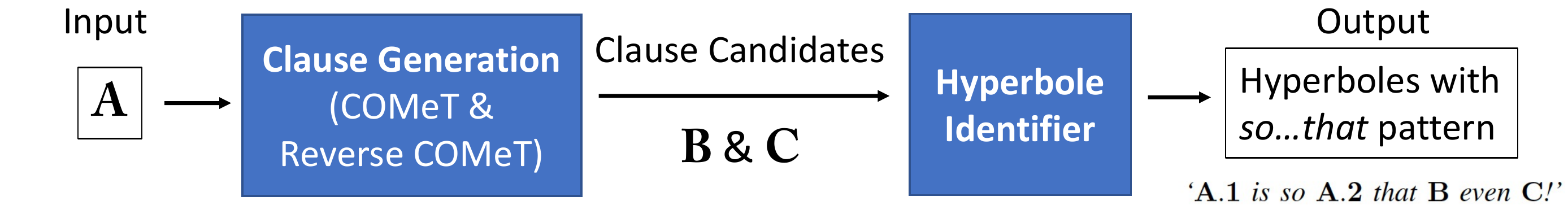}
\caption{A high-level diagram of our hyperbole generation pipeline: \textbf{HypoGen}. We first generate clause candidates with the COMeT and reverse COMeT model, and then rank the candidates with a hyperbole classifier. 
}
\label{fig:diagram}
\vspace*{-3mm}
\end{figure*}

Finally,  we discover that the literal prompt ($\mathbf{A}$) can be further divided into $\mathbf{A.1}$: the subject and $\mathbf{A.2}$: the head word modified by \textit{so} (usually an adjective or adverb). In total, there are three logical relationships between $\mathbf{A B}$: \textbf{1)} $\mathbf{B}$ is related to $\mathbf{A.1}$ (rule 1\&4), \textbf{2)} $\mathbf{B}$ is related to or shares the same attribute with $\mathbf{A.2}$ (rule 2\&5), and \textbf{3)} $\mathbf{B}$ is identical to $\mathbf{A.1}$(rule 3\&6). For example, for \textit{{\color{blue}`He is so tall} that {\color{red}a mountain} {\color[HTML]{998403}looks up to him.}'}, `He' is $\mathbf{A.1}$ and `tall' is $\mathbf{A.2}$.  Since a mountain ($\mathbf{B}$) has the attribute of tall ($\mathbf{A.2}$), but is not capable of looking up ($\mathbf{C}$), this hyperbole a sample from rule 2.

For all six rules, we use Spearman's correlation to measure the inter-annotator agreement (IAA). The IAA score is 0.88, meaning that the raters have substantially strong agreement. We call the annotated data HYPO-so.

\section{Methodology}\label{sec:model}

In this section, we introduce several components for our generation model. In Section \ref{sec:4.1}, we introduce the COMeT model \cite{bosselut2019comet} and its reverse model that favors less frequent and more creative outputs. In Section \ref{sec:model-generation} we design an algorithm to generate multiple hyperbole candidates. Section \ref{sec:rank} explores two possible classifiers as hyperbole identifiers to select the best candidates. A diagram is shown in Figure \ref{fig:diagram}. Furthermore, we propose to use paraphrasing techniques to break the pattern restriction and generate hyperboles with diverse syntactic structures in Section \ref{sec:para}.

\subsection{COMeT and Reverse COMeT Model}\label{sec:4.1}

\paragraph{COMeT and ConceptNet} COMET \cite{bosselut2019comet} is a pre-trained generative model fine-tuned on ConceptNet \cite{speer2017conceptnet}, a knowledge graph of commonsense knowledge in the format of <Entity1 (E1),  Relation (R), Entity2 (E2)>. We utilize the pretrained COMeT model\footnote{\url{https://github.com/atcbosselut/comet-commonsense}} to generate multiple candidates with E1 and R as inputs. For example, given E1 as `the party is lit' and R as `cause desire',
COMET predicts E2 as `to dance'.

\paragraph{Reverse COMeT Model} Now that we have  the  COMeT  model  to  generate  diverse commonsense descriptions from left to right, we also need another model to predict E1 from E2 and R. To this end, we train a reverse COMeT model that takes E2 as input, and E1 as output. That is to say, the ordering of the original ConceptNet tuple is reversed with respect to the COMeT model. 

On top of this, we add two mechanisms to generate even more creative descriptions. First, the reverse COMeT model favors phrases with novel or less frequent words. During the decoding step, we re-score and rank the generated beams. Inspired by mutual information, the re-ranking function is:


\begin{equation}
\vspace*{-2mm}
\mathcal{R_k}=\frac{e^{\frac{P(b_k)}{T}}}{\frac{ \sum_{i=1}^{T}P_{b_k}(i)}{T}},
\label{eq:beam_rescorer}
\end{equation}

\noindent where $P(b_k)$ is
the probability of generation beam $k$, $T$ is the length of beam $k$, $P_{b_k}(i)$ is the unigram probability of the $i^{th}$ word in beam $k$ and $\frac{ \sum_{i=1}^{T}P_{b_k}(i)}{T}$ is the unigram probability that the beam exists in the corpora.
\begin{table}[t!]
\small
\centering
\begin{tabular}{|l|l|}
\hline
\textbf{Retrieved Simile} & \textbf{Created Triplet} \\ \hline
as impertinent as the drama & <drama, HP, impertinent> \\ \hline
as silent as the grave & <grave, HP, silent> \\ \hline
as pale as a sheet & <sheet, HP, pale> \\ \hline
as effortless as breathing & <breathing, HP, effortless> \\ \hline
\end{tabular}
\caption{Examples of the similes we retrieved, and the triplets we created in the format of: <Entity1, Has Property (HP), Entity2)>.}
\label{Table:simile}
\end{table}

Second, we augment the original training triplets in the ConceptNet data\cite{speer2017conceptnet} with figurative samples retrieved from the simile corpora \cite{chakrabarty2020generating}. Table \ref{Table:simile} shows a few examples for the original similes and their relationships. For instance, we map the simile \textit{`as impertinent as the drama'},  to <drama, HasProperty, impertinent>.\footnote{The total number of additional figurative triplets for training the reverse COMeT model is 13,640.}

\subsection{Clause Candidate Generation} \label{sec:model-generation}
\begin{figure}[t!]
\centering
\includegraphics[width=0.3\textwidth]{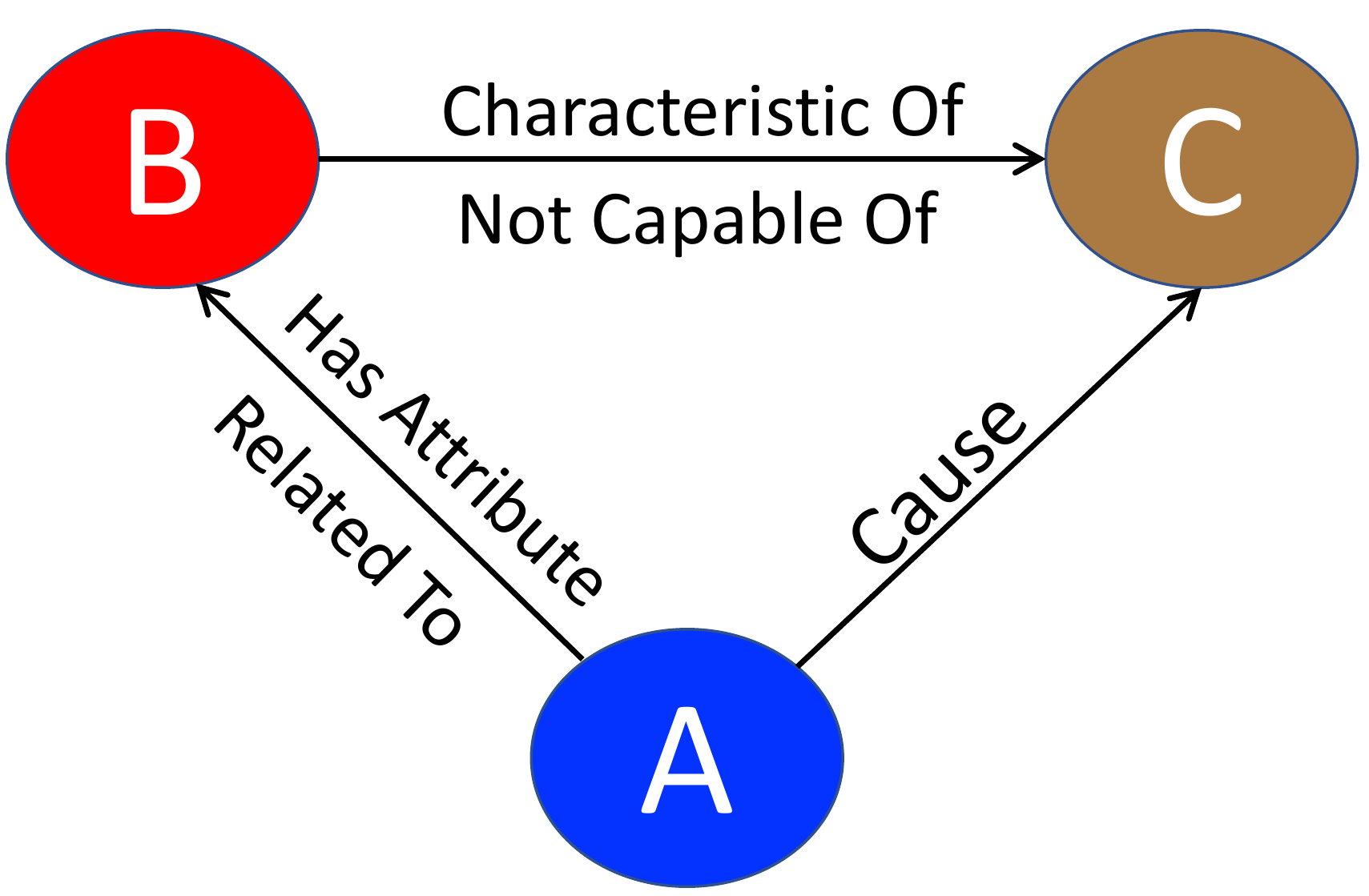}
\caption{An illustration of the generation flow.}
\label{fig:gen_graph}
\vspace*{-3mm}
\end{figure}

Since counterfactuality is a salient component of successful hyperboles, language models view hyperboles as less predictable than literals \cite{troiano2018computational}. Therefore, instead of generating the clause fully, we \textbf{separately} generate the clause's subject ($\mathbf{B}$) and predicate ($\mathbf{C}$). Our generation flow is illustrated in Figure \ref{fig:gen_graph} and Algorithm \ref{algo:1}. 

\begin{algorithm}[t!]
   \caption{Hyperbole Clause Generation}
   \small
    \begin{algorithmic}[1]

      \Function{GenHyper}{$A$}\\
      \textbf{Input: }{Input prompt $\mathbf{A}$ }\\
      \textbf{Output: }{ List of candidate <$\mathbf{B}$, $\mathbf{C}$> pairs $cand$}

        \State Initialize $Bs$, $cand$ to empty list
        \State $subject$, $head\_word$ = parse($\mathbf{A}$) 
        \State $Bs$ += getPreds($subject$, `RelatedTo')
        \State $Bs$ += getPreds($head\_word$, `HasProperty')
        \State $Bs$ += subject
        
        \For{$\mathbf{B}$ in $Bs$}
            \For{$\mathbf{C}$ in getPreds($\mathbf{A}$, `Causal')}
            \State $cand$.append(<$\mathbf{B}$,$\mathbf{C}$>)
            \EndFor
            \For{$\mathbf{C}$ in getPreds($\mathbf{B}$, `CharacteristicOf')}
            \State $cand$.append(<$\mathbf{B}$,$\mathbf{C}$>)
            \EndFor
        \EndFor
        \Return{$cand$}
       \Comment{Fit into the \textit{so...that} pattern.}
       \EndFunction

\end{algorithmic}
\label{algo:1}
\vspace*{-1mm}
\end{algorithm}
\vspace*{-1mm}

\paragraph{Generating $\mathbf{B}$ from $\mathbf{A}$}  We first parse the input prompt ($\mathbf{A}$) into the subject ($\mathbf{A.1}$) and the headword ($\mathbf{A.2}$). We then generate $\mathbf{B}$ using the \texttt{RelatedTo} and \texttt{HasProperty} with the COMeT and reverse COMeT model. Following the COMeT paper \cite{bosselut2019comet}, we also compute the conditional log-likelihood of predicting the object tokens $X$:
\begin{equation}
\vspace*{-2mm}
\mathcal{L}=-\sum_{t=|e1|+|r|}^{|e1|+|r|+|e2|} \log P\left(x_{t} \mid x_{<t}\right),
\label{eq:1}
\end{equation}
where $|e1|$, $|r|$, and $|e2|$ are the number of tokens in e1, relation, and e2, respectively. We denote the likelihood $\mathcal{L}$ as $l_{AB}$ when the likelihood is calculated from generating $\mathbf{B}$ from $\mathbf{A}$.

\paragraph{Generating $\mathbf{C}$ from $\mathbf{A}$ and from $\mathbf{B}$} There are two ways to generate $\mathbf{C}$: from $\mathbf{A}$ and from $\mathbf{B}$. Given $\mathbf{A}$, we can leverage several causal relationships, such as \texttt{CauseDesire}, \texttt{Causes}, and \texttt{HasSubevent}. Given $\mathbf{B}$, we produce i) predicates that $\mathbf{B}$ is not capable of, using \texttt{NotCapableOf} directly available in ConceptNet; and ii) characteristic actions of $\mathbf{B}$, from the following relationships  \texttt{DefinedAs}, \texttt{CapableOf}, \texttt{IsA}, and \texttt{UsedFor}. We also compute the conditional log-likelihoods and call them $l_{AC}$ and $l_{BC}$.

Finally, we assemble pieces of $\mathbf{A}$,  $\mathbf{B}$ and  $\mathbf{C}$ into the `so...that' pattern. The candidate sentence is:  \textit{`$\mathbf{A.1}$ is so $\mathbf{A.2}$ that $\mathbf{B}$ even $\mathbf{C}$!'}.

\paragraph{Grammar Error Correction} When we assemble pieces of $\mathbf{A}$,  $\mathbf{B}$ and  $\mathbf{C}$ into the `so...that' pattern, such manipulation can cause certain grammar errors such as mismatch of verb tenses, or singularity/plurality. While writing a rule-based grammar error correction (GEC) algorithm can be effective for a set of these common errors, we hope to fix open-ended grammar errors. Therefore, we choose the GEC model by \citet{zhao2019improving}, a widely used neural architecture for the GEC problem with copy-augmented architecture and token-level and sentence-level multi-task learning.

\subsection{Hyperbole Candidate Ranking}\label{sec:rank}
We build two classifiers to score and rank the hyperbole candidates. We later compare their performance through human evaluation and ablation in Section \ref{sec:result} and Section \ref{sec:role}.

\paragraph{The Generic Classifier} First, we train a generic hyperbole classification model by finetuning BERT \cite{devlin2018bert} with the data collected in Section \ref{sec:data1}. Before training, we deliberately remove all the keywords such as \textit{I swear}, \textit{literally}, \textit{so … that} to eliminate the influence of superficial cues. We call the model $Clf_G$ and predicted probability $p_{G}$.We call the generation method with $Clf_G$ as classifier \textbf{HypoGen}$_{Gene}$.

\paragraph{The Specific Classifier} The second classifier is specifically designed for hyperboles with the \textit{so...that} pattern. We posit that values of $l_{AB}$, $l_{AC}$, and $l_{BC}$ indicate the intensity of a hyperbole when $Clf_G$ is not fully reliable. Hence, we compute the values of $p_G$, $l_{AB}$, $l_{AC}$ and $l_{BC}$ for 600 \textit{so...that} sentences (half of them are hyperboles and half are literals), and then train a multiple layer perceptron with these four variables as input features. We call the model $Clf_S$ and predicted probability $p_S$: 
\begin{equation}
\vspace*{-2mm}
    p_S = \mathbf{MLP}( p_G, l_{AB}, l_{AC}, l_{BC})
\end{equation}

\noindent Note that to avoid information leakage, the training data for $Clf_G$ and $Clf_S$ do not overlap. We call the generation method with $Clf_S$ as classifier \textbf{HypoGen}$_{Spec}$.


\subsection{Breaking the \textit{so...that} Pattern}\label{sec:para}
So far we have managed to generate hyperboles with the \textit{so...that} pattern. As an extension to our proposed \textbf{HypoGen}, we posit that a paraphrasing module is helpful to break such pattern and hence generate hyperboles with diverse syntactic structures. Specifically, we use the syntactically-controlled paraphrasing model by \citet{sun2021aesop} as an off-the-shelf tool, because it achieves state-of-the-art performances on semantic preservation and syntactic conformation.  It leverages  pretrained  BART \cite{lewis2019bart} and  adds deliberately  chosen  syntactical  control  via  a retrieval-based  selection  module  to  generate fluent paraphrases.

We use \textbf{HypoPara} to denote \textbf{HypoGen}$_{Spec}$ added by such a paraphrasing model.
\section{Experiments}

\subsection{Hyperbole Detection Model}\label{sec:exp-dectect}
\paragraph{$Clf_G$.} Recall that to further remove the influence of superficial clues for hyperboles, we delete all keywords used to crawl hyperboles from Reddit. Next, we balance the training data and then finetune the BERT-base model \cite{devlin2018bert} to train a binary classification model.
We also compare our classification model with that of \citet{troiano2018computational} by testing on their dataset, HYPO-en.

\paragraph{$Clf_S$.} We train a simple MLP for $Clf_S$ and use grid search to find the best hyper-parameters. The best neural network has 2 hidden layers with sizes of 8 and 4. Alpha is \num{1e-4} for regularization.

 \subsection{Baselines}
 
 \paragraph{Sim Retrieval} We first try a naive phrase matching model where we retrieve sentences that contain the input prompt ($\mathbf{A}$).  However, the success rate of exact match is only 3\%, so we utilized a less stringent matching function called Sim Retrieval. Sim Retrieval uses cosine similarity of token embeddings to find the sentence that is semantically similar to a input prompt ($\mathbf{A}$).  For both retrieval based baselines, we retrieve from news commentaries dataset from 2007 to 2020 \footnote{http://data.statmt.org/news-crawl/en/} because the corpus is large and is likely to contain hyperboles. 
 
 \paragraph{Fine-tuned BART} We finetune the model with the input prompts ($\mathbf{A}$) as input to the encoder and the full hyperboles as the output by the decoder.
 
 \paragraph{Ablations of \textbf{HypoGen}} To study the role of each model component, we compare four variations of our main model. We rank the generated hyperbole candidates with 1) $p_G$  (\textbf{HypoGen}$_{Gene}$), 2) $p_S$ (\textbf{HypoGen}$_{Spec}$), 3) $p_G$ and $l_{AC}$ (we call \textbf{HypoGen}$_{Spec}$ w/o $\mathbf{B}$), 4) $p_G$ and $l_{AB}$ (we call \textbf{HypoGen}$_{Spec}$ w/o $\mathbf{C}$).

 \subsection{Evaluation}
 \paragraph{Automatic Evaluation} For creative generation, it is uncommon to have significant n-gram overlap between the machine-generated and the gold-standard sentences. Therefore, instead of BLEU, we use BERTScore \cite{zhang2019bertscore} to measure the semantic similarity between machine outputs and human-written hyperboles. In addition, \citet{troiano2018computational} propose \textit{unexpectedness} to assess the quality of hyperboles, which refers to the fact that hyperboles are less predictable expressions than literals both for humans and language models. We follow their procedure and compute the sentence \textit{expectedness} as its average token probability predicted by GPT2-large \cite{radford2019language}.
 
 \paragraph{Human Evaluation}
 Currently available automatic metrics cannot fully reflect the quality of generated hyperboles. Hence, we also conduct human-based evaluation.
 We first ask the annotators to evaluate if a given sentence is hyperbole, and compute the success rate of each generation model. We then ask a set of 5 criteria to evaluate the generated output: \textbf{1)} Intensity of the hyperbole: extent of the exaggeration, \textbf{2)} Coherency of the hyperbole: how well the clause is reasonably, meaningfully and understandingly related to the prompt, \textbf{3)} Funniness, \textbf{4)} Creativity and novelty, and \textbf{5)} Grammaticality. Each generation is annotated by four human annotators. They are asked to score each criteria  on a scale from 1 (not at all) to 5 (extremely). We evaluate 120 sentences for the gold standard (human) model and each baseline. 
\section{Results}\label{sec:result}
\begin{table}[t!]
\centering
\small
\begin{tabular}{|c|c|c|c|}

\hline
Model    & P    & R    & F-1  \\ \hline
$Clf_G$  & 0.84 & 0.83 & 0.84 \\ \hline
Hype-Par \cite{troiano2018computational} & 0.76 & 0.76 & 0.76 \\ \hline
\end{tabular}
\caption{Performance of $Clf_G$ and the baseline model Hype-Par on the HYPO-en testset \cite{troiano2018computational}}
\label{table:clf1}
\end{table}

\pgfplotstableread[row sep=\\,col sep=&]{
    interval & carT \\
    $Clf_G$     &  83.9    \\
    $+l_{AB}$     &  85.1  \\
    $+l_{AC}$    &  86.4  \\
    $+l_{BC}$   &  85.5  \\
    $Clf_S$   &  87.4 \\
    }{\mydata}
    
\begin{figure}
\centering
\small
\begin{tikzpicture}
\centering
\small
   \begin{axis}
        [ticks=none,
        ybar,
        ymin = 83.5,
            width=7cm,
            height=5.5cm,
            symbolic x coords={$Clf_G$,$+l_{AB}$,$+l_{AC}$,$+l_{BC}$, $Clf_S$ },
            xtick=data,
            nodes near coords,
            legend style={at={(0.45,0.99)},anchor=north east}]
        \addplot table[x=interval,y=carT]{\mydata};
        \legend{Accuracy(\%)}
    \end{axis}
\end{tikzpicture}
\vspace*{-3mm}
\caption{Performance of two hyperbole classifiers ($Clf_G$ and $Clf_S$) on \textit{so...that} patterns. We also show ablations of each variable:  $l_{AB}$, $l_{AC}$ and $l_{BC}$.}\label{Figure:histogram}
\end{figure}
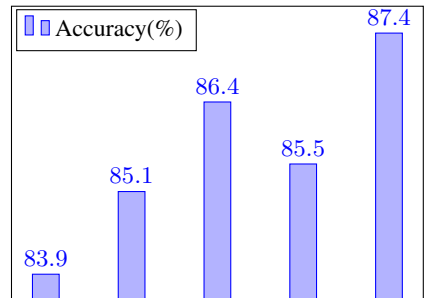

\subsection{Performance of the Classification Model}


\begin{table*}[t!]
\small
\centering
\begin{tabular}{|l|l|c|c|c|c|c|c|}
\hline
\multicolumn{2}{|l|}{Model}                    & Success Rate                        & Intensity                         & Coherency                         & Funniness                         & Creativity                        & Grammar                           \\ \hline
\multicolumn{2}{|l|}{Human}                   & \textbf{84.2\%}                     & \textbf{3.50}                     & \textbf{3.41}                     & \textbf{2.81}                     & \textbf{3.05}                     & \textbf{3.82}                     \\ \hline
\multirow{3}{*}{Baselines} &  Naive Retrieval      & 3.0\%                               & /                                 & /                                 & /                                 & /                                 & /                                 \\ 
& Sim Retrieve & 28.9\%                              & 2.51                              & 2.56                              & 2.14                              & 2.36                              & 2.78                              \\ 
& Fine-tuned BART                    & 44.6\%                              & 2.65                              & 2.78                              & 2.23                              & 2.71                              & {\ul{3.27}$^\dagger$} \\ \hline
\multirow{5}{*}{Proposed} & \textbf{HypoGen}$_{Gene}$                 &  64.1\%      & 3.03                              & {\ul{2.89}$^\dagger$} & 2.46                              & 2.84                              & 3.20                              \\ 
& \textbf{HypoGen}$_{Spec}$    w/o $\mathbf{B}$     & 65.2\%                              & 3.12                              & 2.86                              & 2.44                              & 2.80                              & 3.19                              \\ 
& \textbf{HypoGen}$_{Spec}$     w/o $\mathbf{C}$     & 66.3\%                              & 3.19                              & 2.79                              & 2.50                              & 2.89                              & 3.12                              \\ 
& \textbf{HypoGen}$_{Spec}$             & {\ul{67.8\%}$^\dagger$} & {\ul{3.23}$^\dagger$} & 2.85                              & { \ul{2.54}$^\dagger$} & {\ul{2.98}$^\dagger$} & 3.13                              \\ \cline{2-8}
& \textbf{HypoPara}          & 48.0\%                              & 3.17                              & 2.81                              & 2.40                              & 2.75                              & 3.17                              \\ \hline
\end{tabular}
\caption{Human evaluation results on the success rate and five criteria of hyperbole quality: intensity, coherency, funniness, creativity or novelty, and grammarcality. Boldface in black denotes the human performance; underscore with $\dagger$ denotes the best performance among models.}
\label{table:main-result}
\vspace*{-3mm}
\end{table*}

\begin{table}[t!]
\centering
\small
\begin{tabular}{|@{ }l|c@{\ \ }|@{\ \ }c@{\ \ }|@{\ \ }c|c|}
\hline
\multicolumn{1}{|@{ }l|}{\multirow{2}{*}{Model}}                      & \multicolumn{3}{c|}{BERTScore}                                                                              &                                    \\ \cline{2-4}
 & P                                  & R                                  & F1                                & \multirow{-2}{*}{\begin{tabular}[c]{@{}c@{}}Expect-\\edness\end{tabular}}           \\ \hline
Human                                         & \textbf{1.00}                     & \textbf{1.00}                     & \textbf{1.00}                     & \textbf{0.095}                     \\ 
Sim Retrieval                                 & 0.19                              & 0.28                              & 0.23                              & 0.139                              \\ 
Fine-tuned BART                                 & 0.24                              & 0.29                              & 0.27                              & 0.115                              \\ \hline
\textbf{HypoGen}$_{Gene}$                                      & 0.31                              & 0.29                              & 0.30                              & 0.087                              \\ 
\textbf{HypoGen}$_{Spec}$ w/o B                              & 0.31                              & {\ul{0.29}$^\dagger$} & {\ul{0.30}$^\dagger$} & 0.084                              \\ 
\textbf{HypoGen}$_{Spec}$ w/o C                             & 0.29                              & 0.27                              & 0.28                              & 0.083 \\ 

\textbf{HypoGen}$_{Spec}$                            & {\ul{0.31}$^\dagger$} & 0.29                              &  0.30 & {\ul{ 0.083}$^\dagger$} \\ \hline
\textbf{HypoPara}                                       & 0.30                              & 0.27                              & 0.28                              & 0.093                              \\ \hline

\end{tabular}
\caption{Automatic evaluation results of our model \textbf{HypoGen} and baselines. We report the precision, recall and F1 of BERTScore (higher is better), and expectedness (lower is better). Boldface in black denotes the human performance; underscore with $\dagger$ denotes the best performance among models.}
\label{table:auto-eval}
\vspace*{-3mm}
\end{table}

\paragraph{The Generic Classifier.} Table \ref{table:clf1} shows the accuracy of $Clf_G$ and the previous SOTA Hype-Par \cite{troiano2018computational} that uses Skip-Gram representations and several manually defined features.  
Even though $Clf_G$ is trained on HYPO-Red and tested on HYPO-en (Hype-Par is trained and tested on the same dataset, HYPO-en), our $Clf_G$ still outperforms Hyper-Par by 8\%. Tested on HYPO-Red, $Clf_G$ achieves a score of 83.35\%. We cannot see how well Hype-Par does on HYPO-Red, because Hype-Par requires computing hand-crafted features on the training data, which is not publicly available. 

\paragraph{The Specific Classifier.} Figure \ref{Figure:histogram} reports the performances of $Clf_G$ and $Clf_S$ on the task of identifying hyperboles containing \textit{so...that} patterns. $Clf_G$ alone already achieves satisfactory accuracy (83.9\%), and $Clf_S$ is 3.5\% better than $Clf_G$. With the addition of $l_{AB}$, $l_{AC}$ or $l_{BC}$, model performances have increased by 1.2\%, 2.5\%, or 1.6\%. Among them, the causal relation between A and C contributes most.

\begin{table*}[t!]
\small
\centering
\begin{tabular}{|p{1.8cm}||p{9.6cm}|p{0.6cm}|p{0.6cm}|p{0.6cm}|p{0.6cm}|}
\hline

\textbf{System}                                                          & \textbf{Generation}                                                                                                               & \textbf{Int}          & \textbf{Coh}          & \textbf{Fun}          & \textbf{Cre}          \\ \hline
\end{tabular}

\begin{tabular}{|p{1.8cm}||p{9.6cm}|p{0.6cm}|p{0.6cm}|p{0.6cm}|p{0.6cm}|}
\hline
Human                                                           & When the party is so   lit even the wardrobe is dancing!                                                                 & \textbf{4.50} & \textbf{4.50} & \textbf{3.75} & \textbf{4.25} \\ 
Sim Retrieval & The party is   so lit that happened after was crazy I thought I was gonna die! & 3.25          & 2.75          & 2.00          & 2.00          \\ 
 BART       & The party is so lit   that even the bugs had too give it a yelp review!                                                  & 3.75          & 2.75          & 2.50          & {\ul 3.50}    \\ 
HypoGen 1                                                     & The party is so lit   that even the street wants to have fun!                                                            & 3.75          & 3.25          & 2.50          & 2.75          \\ 
HypoGen 2                                                     & The party is so lit   that even the city gets drunk!                                                                     & {\ul 4.00}    & {\ul 3.75}    & {\ul 3.00}    & 3.00          \\ 
HypoPara 1                                                    & What a lit party that   the street wants to have fun with it!                                                            & 3.50          & 2.75          & 2.25          & 2.75          \\
HypoPara 2                                                    & Why is party so lit   that the city is drunk?                                                                            & 4.00          & 3.50          & 2.50          & 3.00          \\ \hline
\end{tabular}

\begin{tabular}{|p{1.8cm}||p{9.6cm}|p{0.6cm}|p{0.6cm}|p{0.6cm}|p{0.6cm}|}
\hline
Human         & His drawing is so   bright that I cannot open my eye!                                                                                  & 4.50                    & {\ul 4.75}             & 2.75                   & {\ul 4.50}              \\ 
Sim Retrieval & \begin{tabular}[c]{@{}l@{}}His drawing is so   bright, at first its discoverers thought something was wrong,\end{tabular} & 2.25          & 3.00             & 2.75             & 4.25             \\ 
BART          & His drawing is so   bright even god gave up with you before giving him a chin.                                                         & 3.25                   & 2.50                    & 2.75                   & 3.00                      \\ 
HypoGen 1   & His drawing is so   bright that even sun adjusts focus!                                                                                & \textbf{5.00}             & {\ul 4.75}             & {\ul 4.25}             & 4.00              \\ 
HypoGen 2   & His drawing is so   bright that even stars fade away!                                                                                  & {\ul 4.75}             & \textbf{5.00}             & \textbf{4.50}           & \textbf{4.75}          \\
HypoPara 1  & How can I learn about   such a bright drawing when the sun adjusts the focus?                                                          & 2.50                    & 2.75                   & 2.25                   & 2.50                    \\
HypoPara 2  & I 'm sure his   picture's so bright that the stars are gone.                                                                           & 4.00                      & 4.00                      & 3.50                    & 3.75                   \\\hline
\end{tabular}

\begin{tabular}{|p{1.8cm}||p{9.6cm}|p{0.6cm}|p{0.6cm}|p{0.6cm}|p{0.6cm}|}
\hline
Human         & Your forehead is so   big even a 787 can land on it.                                                     & \textbf{5.00}             & \textbf{4.75}          & \textbf{4.75}          & \textbf{4.75}          \\ 
Sim Retrieval & Your forehead is so   big that ordinarily would have threatened to ruin a perfect day for watching   TV. & /          & /             & /             & /             \\
BART          & Your forehead is so  big that even your hairline is running away from it.                             & 4.25                   & \textbf{4.75}          & {\ul 4.00}                & 4.00                      \\ 
HypoGen 1   & Your forehead is so   big even Eiffel Tower can not fit inside of your head                              & 4.25                   & 3.25                   & 3.75                   & 3.75                   \\
HypoGen 2   & Your forehead is so   big even universe wants to inhabit!                                                & {\ul 4.75}             & 3.25                   & {\ul 4.00}                & {\ul 4.25}             \\
HypoPara 1  & Does eiffel tower fit   in your head?                                                                   & 3.75                   & 3.00                      & 3.50                    & 3.75                   \\ 
HypoPara 2  & You have such a big   forehead that even the universe would want to inhabit it.                           & 4.50                    & {\ul 4.50}              & 4.00                      & 4.00                      \\ \hline
\end{tabular}

\begin{tabular}{|p{1.8cm}||p{9.6cm}|p{0.6cm}|p{0.6cm}|p{0.6cm}|p{0.6cm}|}
\hline
Human         & The young artist is   so productive, even paintings get moved and start to paint themselves!                               & \textbf{4.25} & \textbf{4.75} & \textbf{4.00} & \textbf{3.75} \\ 
Sim Retrieval & The young artist is   so productive that age and I didn't make the same mistakes because I was able   to learn from her's. & /          & /             & /             & /    \\ 
BART          & The young artist is   so productive that even Shia Labeouf tells you not to do it.                                         & 3.00          & 2.25          & {\ul 3.00}    & 2.50          \\ 
HypoGen 1   & The young artist is so productive that Botticelli removes paint from his wall!                                 & {\ul 4.00}    & 3.00          & 2.75          & {\ul 3.25}    \\ 
HypoGen 2   & The young artist is so productive that Botticelli wants to retire!                                                 & 3.75          & 3.50          & 2.75          & 2.75          \\ 
HypoPara 1  & Will give rise to the art of youth and even stop selling Botticelli's paintings!                                   & 3.75          & {\ul 3.25}    & 2.25          & 2.50          \\ 
HypoPara 2  & What is the success   of young artists for letting Botticelli retire?                                                      & 2.75          & 2.75          & 2.00          & 2.75          \\ \hline
\end{tabular}

\caption{Examples of generated outputs from human and different models, and their intensity, coherency, funniness, and creativity scores. We
show average scores (over four annotators) on a 1-5 scale, with 1 denoting the worst and 5 the best. The
boldface numbers denote the best scores, and underlined numbers denote the second best scores. HypoGen 1 and HypoGen 2 represent two hyperboles generated by \textbf{HypoGen}$_{Spec}$}
\label{table:examples}
\vspace*{-4mm}
\end{table*}

\subsection{Evaluation Results}
We report the results of human and automatic evaluation in Table \ref{table:main-result} and Table \ref{table:auto-eval}.
\paragraph{Automatic Evaluation}

Table \ref{table:auto-eval} shows the precision, recall, and F1 of BERTScore and the \textit{expectedness} value of our systems and the baselines. Compared with the baselines,  \textbf{HypoGen}$_{Spec}$ achieves high BERTScore, meaning that the generations of \textbf{HypoGen}$_{Spec}$ are semantically similar to human-written hyperboles. For \textit{expectedness} scores, the retrieval method and fine-tuned BART tend generate more `normal' and predictable outputs than our systems. However, \textbf{HypoGen} is even less predictable than human-written hyperboles. A possible explanation is that human-written ones are both coherent and exaggerating, containing more conjunction words (e.g., the, and, so, that) which contribute to the average word probability.

\paragraph{Human Evaluation}

Table \ref{table:main-result} reports the scores of the five human-evaluated criteria for our model and its variable, human written hyperboles, and the baseline models. To better understand the merits of our proposed model, we also provide four examples of the generated outputs in Table \ref{table:examples}. It is interesting that 
\textbf{HypoGen}$_{Spec}$ is annotated to achieve creativity close to that of humans. We attribute such a high creativity score to the counterfactuality introduced in Section \ref{sec:model-generation}. 

For all automatic generation methods, \textbf{HypoGen}$_{Spec}$  has the highest success rate (67.8\%), intensity of hyperbole (3.23/5), funniness (2.54/5) and creativity (2.98/5). On the other hand, the BART model is the best at producing grammatically correct outputs. Even with the grammar-error-correction model provided by \citet{zhao2019improving}, \textbf{HypoGen} still suffers from grammar errors.

\subsection{Breaking the \textit{so...that} Pattern}\label{sec:compare}
Based on the evaluation results in Table \ref{table:main-result} and the examples in Table \ref{table:examples}, it is clear that we are able to generate hyperboles with diverse syntactic structures through paraphrasing. However, the success rate and quality of hyperboles become lower.  We believe that since \textbf{HypoGen} and \textbf{HypoPara} each has its own benefits, a trade-off between diversity and intensity is inevitable. Moreover, since we leverage off-the-shelf paraphrasing models, we believe the performance of \textbf{HypoPara} will improve with the development of paraphrasing techniques.

\section{Role of Each Component}\label{sec:role}
Here we analyze the role of $\mathbf{A}$, $\mathbf{B}$, and $\mathbf{C}$ in \textbf{HypoGen}. Ablations of our own models are colored in the grey background in Table \ref{table:main-result}.  First, we discover that \textbf{HypoGen}$_{Gene}$ is better at selecting coherent and grammar correct generations then \textbf{HypoGen}$_{Spec}$. A possible explanation is that \textbf{HypoGen}$_{Gene}$ is finetuned on BERT, and that pretrained language models are good at selecting coherent text. However, \textbf{HypoGen}$_{Spec}$ is still considered the best model, because it has the highest success rate and generate the most exaggerated, fun, and creative hyperboles. 

Second, compared with the predicate ($\mathbf{C}$), the subject of clause ($\mathbf{B}$) contributes more to the funniness score and creativity score. We posit that the interplay between $\mathbf{A}$ and $\mathbf{B}$ (and also between $\mathbf{B}$ and $\mathbf{C}$) is the dominant factor of novelty, funniness and creativity. Similarly, the predicate ($\mathbf{C}$) which is responsible as a result of input, contributes more to the coherency score. We hence posit that the interplay between $\mathbf{A}$ and $\mathbf{C}$ determines how well our generation is reasonable and understood.

\section{Related Work}
\subsection{Linguistic Studies on Hyperboles}
Our generation model is partially inspired and back-boned by various linguistic studies about hyperboles. \citet{claridge2010hyperbole} classify hyperboles into word/phrase level and clause/sentence level. The former can be easily achieved via lexicon substitution \citep{norrick2012semantics}, while the latter requires more sophisticated world knowledge and hence is more creative, interactive and challenging. 

\citet{mccarthy2004there, mora2009all,claridge2010hyperbole} identify hyperbole as the creation of impossible worlds, unchallenged
counterfactuality and syntactic support.
\citet{kunneman2015signaling} focus on the presence of language intensity as a potential cue to hyperbole. \cite{backlund1973collocation, lorenz2002really} further study the \textit{so} + \textit{(adj/adv)} + \textit{that} + \textit{a declarative clause} as a significant intensification pattern that has both prototypical syntactic and semantic function as overstatement.

\citet{claridge2010hyperbole} find out that in the \textit{so...that} pattern, the content clauses always express the resultant meaning of the prompts and that the clauses itself creates impossible worlds. Such discoveries motivate us to comprehensively uncover the sensical (commonsense or counterfactuality) relationships behind hyperboles in Section \ref{sec:rel-anno}.

\subsection{Hyperbole Detection}
\citet{troiano2018computational} and \citet{kong2020empirical} explore statistical and neural based approaches to automatic hyperbole detection in English (HYPO-en) and Chinese (HYPO-cn) corpora.  \citet{troiano2018computational} introduce hand-crafted features while \citet{kong2020empirical} achieve better performance by jointly training with such hand-crafted features and a directional skipgram. We also train a hyperbole identifier as part of the generation model. However, for our classifier, we finetune the BERT model.

\subsection{Figurative Generation}
Recent years have witnessed increased interest in creative and figurative language generation. \citet{yu2019avoid} generate metaphor unsupervisedly by extracting the metaphorically-used verbs; \citet{chakrabarty2021mermaid} propose a metaphor generation method with symbolism and discriminative decoding; \citet{stowe2021metaphor} study diverse metaphor generation using conceptual mapping. Given a pair of homophones, \citet{yu2018neural} train a conditional neural language model with an decoding algorithm for pun generation;
\citet{he2019pun} tackle the same task with a local-global surprisal principle and a retrieve-and-edit pipeline; \citet{luo2019pun} on the other hand propose an adversarial pun generative network.

Generating hyperboles or exaggerations is a new task. To the best of our knowledge, we are the first to work on hyperbole generation. The closest work is that of \citet{chakrabarty2020generating}, who propose an end-to-end approach for simile generation that also utilizes commonsense knowledge predicted by COMeT \cite{bosselut2019comet}. However, they only utilize the \texttt{PROPERTY} relation to replace certain parts of literal sentences. We leverage a more complex set of commonsense knowledge during the generation time, and target at a different trope of figurative language.

\section{Conclusion and Future Work}
We are the first to tackle the novel task of hyperbole generation at the \textit{clause or sentence level} . We start with the representative \textit{so...that} pattern, partition it into three components and analyze the logical relationships among them. Our proposed model \textbf{HypoGen} first generates commonsense and counterfactual predictions, and then selects top-ranking candidates as hyperboles. Our experimental results show that \textbf{HypoGen}$_{Spec}$ is able to generate hyperboles with high success rate and high semantic intensity, funniness, and creativity scores.

In addition, we propose \textbf{HypoPara} as a diversity-oriented generation approach. Follow-up works on hyperbole generation without relying on any patterns can use \textbf{HypoPara} as a baseline. Both our \textbf{HypoGen} and \textbf{HypoPara} can be applied to downstream applications such as dialog systems and storytelling, to improve their interestingness and engagement.

\section*{Acknowledgement}
We thank the anonymous reviewers and Emily Sheng, Yu (Hope) Hou, and Nuan Wen for their useful feedback. We also appreciate the kind efforts of the undergraduate students from PlusLab to annotate the relations. This work is supported by the Machine Common Sense (MCS) program
under Cooperative Agreement N66001-19-2-4032
with the US Defense Advanced Research Projects
Agency (DARPA). The views and the conclusions
of this paper are those of the authors and do not
reflect the official policy or position of DARPA.

\section*{Ethics Considerations}

We understand and respect user privacy. The HYPO-Red dataset is collected from Reddit totally anonymously, and does not reveal any details about the users' personal information, including name, racial or ethnic origin, religious or philosophical affiliation or beliefs, sexual orientation, etc. 

Our proposed methods are based on the pretrained language model. It is known that pretrained language models could capture the bias reflected in the training data \cite{sheng2019woman,wallace2019universal}. Considering the nature of exaggeration or overstatement, the context and sentiment of the literal input prompt also affect the our generated hyperboles. Therefore, our models may potentially generate offensive content for certain groups or individuals.  We suggest to carefully examine the potential biases before deploying the models to real-world applications.

\bibliography{anthology,custom}
\bibliographystyle{acl_natbib}

\newpage





\end{document}